\documentclass[11pt,a4paper]{article}
\usepackage{authblk}
\usepackage[hyperref]{acl2019}
\usepackage{enumitem}
\usepackage{times}
\usepackage{latexsym}
\usepackage{todonotes}
\usepackage{amsmath}
\usepackage{multirow}
\usepackage{graphics}
\usepackage{url}
\usepackage[linesnumbered, ruled, vlined]{algorithm2e}
\usepackage{enumitem}
\usepackage{subcaption}
\setlist[itemize]{noitemsep, topsep=0pt}
\setlist[enumerate]{noitemsep, topsep=0pt}
\usepackage{varwidth}
\usepackage{cleveref}
\aclfinalcopy 

\title{Understanding the Role of Scene Graphs in Visual Question Answering}

\author[1\thanks{\hspace{2pt} Equal Contribution}]{Vinay Damodaran}
\author[1$^*$]{\hspace{5pt}Sharanya Chakravarthy}
\author[1$^*$]{Akshay Kumar}
\author[1$^*$]{Anjana Umapathy}
\author[1]{\authorcr Teruko Mitamura}
\affil[1]{Carnegie Mellon University \\ \texttt{\{vdamodar, sharanyc, akshayak, aumapath, teruko\}@cs.cmu.edu}} 
\author[2]{Yuta Nakashima}
\author[2]{Noa Garcia}
\affil[2]{Osaka University \\ \texttt{\{n-yuta, noagarcia\}@ids.osaka-u.ac.jp}}
\author[3]{Chenhui Chu}
\affil[3]{Kyoto University \\ \texttt{chu@i.kyoto-u.ac.jp}}

\date{}

\begin{document}
\maketitle
\begin{abstract}

Visual Question Answering (VQA) is of tremendous interest to the research community with important applications such as aiding visually impaired users and image-based search. In this work, we explore the use of scene graphs for solving the VQA task. We conduct experiments on the GQA dataset \citep{hudson2019gqa} which presents a challenging set of questions requiring counting, compositionality and advanced reasoning capability, and provides scene graphs for a large number of images. We adopt image + question architectures for use with scene graphs, evaluate various scene graph generation techniques for unseen images, propose a training curriculum to leverage human annotated and auto-generated scene graphs, and build late fusion architectures to learn from multiple image representations. We present a multi-faceted study into the use of scene graphs for VQA, making this work the first of its kind.
\end{abstract}

\section{Introduction}
The task of Visual Question Answering (VQA) requires a model to answer a free-form natural language question that is based on an image. It is an important Vision+Language (V+L) task that has numerous real-world applications such as image-based search, search and rescue missions, visually capable AI assistants and aiding visually impaired users. A number of VQA datasets such as VQA 2.0 \citep{antol2015vqa}, CLEVR \citep{johnson2017clevr} and GQA \citep{hudson2019gqa} have been introduced, fuelling research in the area. 
 
Research in VQA includes works that find biases in the data, demonstrating the need for improved datasets 
\citep{agrawal2018don}, pre-trained models such as \textsc{lxmert} \citep{tan2019lxmert} and \textsc{uniter} \citep{chen2020uniter} that beat the state of the art on a variety of V+L tasks and models such as neural module networks \citep{andreas2016neural} constructed to heighten interpretability, among many others. However, few works have explored the use of scene graphs for the VQA task. Scene graphs provide a graphical representation of the image, containing information about objects and relationships between them. This representation is more advantageous than the typical object features extracted from an image since it is in natural language and allows for greater interpretability. The recently released GQA dataset \citep{hudson2019gqa} contains 85,638 scene graphs for images in the training and validation sets, spurring research into the use of scene graphs for VQA. 

In this work, we explore a set of research questions related to the use of scene graphs for VQA. Given the success of pre-trained models for other V+L tasks, we evaluate the effectiveness of transfer learning for GQA. The use of scene graphs requires converting them from a graphical representation to one that can be undeerstood by a neural network. We adopt the popular Graph Network model \citep{battaglia2018relational} to produce scene graph encodings and then adopt image-based question answering models to use these encodings. One of the challenges in using scene graphs lies in generating them for unseen images. We compare multiple scene graph generation models and study the impact of scene graph quality on model performance. We devise a novel training curriculum aimed at leveraging both ground truth and generated scene graphs to avoid a mismatch between training and testing scene graphs. Additionally, we explore late-fusion ensembles with the goal of using images as well as scene graphs, and combining the strengths of scene graph use with those of pre-training. We present the first in-depth analysis of the use of scene graphs for VQA. We will release our source code, end-to-end pipelines and pre-trained checkpoints in our GitHub repository\footnote{\url{https://github.com/sharanyarc96/scene-graphs-for-vqa}}.

The remainder of this paper is structured as follows. In Section \ref{sec:related_work} we introduce existing approaches to VQA and formally define our task in Section \ref{sec:task-definition}. We describe our methodology and experimental setup in Section \ref{sec:proposed_approach} and \ref{sec:experimental_setup} respectively. In Section \ref{sec:results}, we provide an in-depth analysis of our results. We conclude the paper with a summary of our contributions in Section \ref{sec:contributions}.

\section{Related Work}
\label{sec:related_work}
In this section we discuss VQA datasets, state of the art approaches for VQA, usage of scene graphs, and methods of scene graph generation.

\subsection{Datasets for VQA}
\citet{antol2015vqa} released the VQA 2.0 dataset containing 204,721 real images and 1,105,904 open-ended questions requiring open-ended or multiple-choice answers. Prior works have identified statistical biases in VQA 2.0 with educated guess approaches working remarkably well. The CLEVR dataset \citep{johnson2017clevr} was introduced with the goal of mitigating these biases and testing the visual reasoning capability of models using a large number of compositional questions. However, the number of objects in CLEVR is severely limited and the number of possible answers is just 28. Moreover, the images are synthetically generated. In this work, we conduct experiments on the GQA dataset \citep{hudson2019gqa} which contains few statistical biases, retains the semantic richness of real images, contains open-ended questions of various degrees of compositionality and has a much larger vocabulary than CLEVR. Importantly for this work, it provides cleaned up versions of Visual Genome scene graphs \citep{krishna2017visual} for images in the training and validation sets.    

\subsection{Approaches to VQA}
With the introduction of the VQA 2.0 and CLEVR datasets, many models have been proposed for the VQA task. This section covers some task specific and pre-trained models, each of which have beaten prior state of the art approaches for VQA. 

\subsubsection{Task Specific Models}
We classify the set of models trained from scratch for a specific task on a single dataset as task specific. While the performance of these models may not match that of models which leverage transfer learning, they present interesting architectures that can be modified to use scene graphs. We discuss two such models here. \\
\textbf{Bottom-Up Top-Down Attention (BUTD):}
\citet{anderson2018bottom} propose a combined bottom-up and top-down mechanism for vision and language representation learning. For VQA, top-down attention uses the question, usually predicting an attention distribution over a uniform grid of equally-sized image regions. In contrast, bottom-up attention proposes a set of salient image regions using Faster R-CNN \citep{ren2016faster}, which can be attended on. This architecture has proven beneficial for VQA and image captioning.
\\
\textbf{MAC:} \citet{hudson2018compositional} introduce a neural network architecture that decomposes a question answering task into a series of attention-based reasoning steps, thus increasing the interpretability of the model's decision making process. Given a knowledge-base $K$ (image) and a task description $Q$ (question), the control unit \textsc{(cu)} of the model identifies a series of operations to be performed, the read unit \textsc{(ru)} extracts information needed to perform the operation from the image, and the write unit \textsc{(wu)} integrates the information into an internal cell state to produce an intermediate result. The \textsc{mac} model is one of the baselines introduced with the GQA dataset. Since, it is not a pre-trained model, it can easily be extended to use different knowledge bases without the additional overhead of pre-training.

\subsubsection{Pre-Trained Models}
In recent times, the introduction of pre-trained models such as BERT \citep{devlin-etal-2019-bert} has advanced the state of the art on a number of Natural Language Processing tasks. Similar benefits of using Transfer Learning have been seen for vision based tasks. Thus, it is not surprising that  pre-training has also had a huge impact on the performance on several Vision + Language tasks. \citet{chen2020uniter} try to learn a UNiversal Image-TExt Representation (\textsc{uniter}) through large-scale pre-training over four image + text datasets. To this end, they design four pre-training tasks:  Masked Language Modeling, Masked Region Modeling, ImageText Matching and Word-Region Alignment. \textsc{uniter} achieves state of the art performance on challenges like VQA 2.0  and NLVR2, beating pre-trained models like \textsc{lxmert} \citep{tan2019lxmert} and task-specific models like \textsc{mac}. 

\subsection{Using Scene Graphs for VQA}
To the best of our knowledge, only a few prior works have used scene graphs for VQA. \citet{hudson2019gqa} report\footnote{\url{https://github.com/stanfordnlp/mac- network/issues/22}} a validation set accuracy of 83.5 using a ``perfect sight" \textsc{mac} which replaces object features with a representation of the scene graph. The exact details of the scene graph encoding are unclear. \citet{gn_mac} introduce another perfect sight model where the encoding of the scene graph is learnt using a Graph Network and is passed in place of the image to \textsc{mac}. Their model obtains a validation accuracy of 96.3 using ground truth scene graphs. \citet{hudson2019learning} propose the task-specific Neural State Machine (NSM) which generates a probabilistic scene graph for the image and performs a series of reasoning operations over it. The generated scene graph contains a probability distribution for predicted objects, attributes, and relations. This allows the model to make predictions despite errors in the graph. While NSM gets a high accuracy, the lack of an open-source implementation and missing details required for reproducibility make comparisons with this model difficult. 

\subsection{Using scene graphs for other tasks}
Scene graphs have proven to be useful in various domains of computer vision and visual understanding. \citet{johnson2018image} studied the role of scene graphs in generating images using natural language descriptions. In the field of image outpainting, the work by \citet{li2019controllable} has shown that the use of scene graphs produces much more effective visual results and quantitative evaluations. \citet{lee2020neural} proposes a unique Neural Design Network that uses scene graphs to great effect in the context of layout generation. The effectiveness of scene graphs displayed in these fields further motivates the need to understand the impact of scene graphs in the domain of visual question answering.

\subsection{Scene Graph Generation}
The task of Scene Graph Generation has become an increasingly sought out challenge after discovering its potential to surpass the limits on visual reasoning and question answering tasks. Most works in scene graph generation rely on pre-trained object feature extractors like Faster-RCNN \citep{ren2016faster} and Mask-RCNN \citep{he2017mask} to capture the object and attribute properties in the image. One of the earlier works in the field of scene graph generation, by \citet{zellers2018neural} first introduced the concept of eliminating bias in the global context of determining relations with regards to relational edges. \citet{xu2017scene} first aimed to solve the task of scene graph generation using standard RNNs which learns to iteratively improve its predictions via message passing. Another top performing model by \citet{tang2019learning} constructs scene graphs in the form of tree structures (using TreeLSTMs) and performs very well on a plethora of visual reasoning tasks. The recent work by \citet{tang2020unbiased} looks to generate unbiased scene graphs with more specific relations across edges by building causal graphs and uses counterfactual causality from the trained graph to infer the effect from bad bias.
\section{Task Definition}
\label{sec:task-definition}
In this work, we tackle the multimodal VQA task which involves answering a question $q$ given an image $I$. The question can be expressed as the sequence $(w_1, w_2, ..., w_n)$ where $w_i$ is a single word/token. We work with the following representations of the image: 
\begin{itemize}[noitemsep,topsep=0pt]
    \item Spatial Feature: A single vector $S$ for the entire image
    \item Object Features: A set of $N^v$ vectors $O = (o_1, o_2, ..., o_{N^v})$ representing $N^v$ objects in the image 
    \item Scene Graph: A graphical representation $g$ of the contents of the image
\end{itemize}

We formally define the scene graph in Equation ~\eqref{eq:sg_def}, where $V$ is the set of nodes representing objects in the image, $E$ is the set of edges representing relations between objects, and $u$ is a global feature shared by all components of the graph. $N^v$ is the number of objects in the image, and $N^e$ is the number of relations. $v_i^{name}$ is the name of the object corresponding to the node $v_i$ and $v_i^{attr}$ is the set of attributes of the object. Similarly,  $e_j^{name}$ is the name of the relation between a source object / node $e_j^s$ and a receiver object $e_j^r$.
\begin{equation}
\label{eq:sg_def}
\begin{gathered}
    g = (E, V, u) \\
   V = (v_1, v_2, ..., v_{N^v}); v_i = (v_i^{name}, v_i^{attr}); \\
       E = (e_1, e_2, ..., e_{N^e}); e_j = (e_j^{name}, e_j^s, e_j^r); \\ e_j^s, e_j^r \in V
\end{gathered} 
\end{equation}

We work in a discriminative setting wherein, given $Q$ and one or more representations of $I$ as input, the model must select an answer for $Q$ from a fixed set of $n$ answers, A = ($A_1$, $A_2$, ..., $A_n$).

\section{Proposed Approach}
\label{sec:proposed_approach}
In this section we present the details of our methodology.
\subsection{Baselines}
\label{sec:baselines}
We define two multimodal baselines for VQA, one using the image spatial features $S$, namely the Concatenation Baseline \textsc{(concat)}, and another using the image object features $O$, namely the Attention Baseline \textsc{(attn)}. For all our baselines, we use mean pooled BERT embeddings $B$ of the question $q$. We also define unimodal baselines.

\subsubsection{Multimodal Baselines}
In the \textsc{concat} model, mean pooled representations of $O$ are passed in parallel with $B$ through two fully-connected layers with ReLU activation to get $b*1024$ vectors for both modalities, where $b$ is the batch size. These vectors are concatenated and passed through two fully-connected layers with ReLU activation in between.

\begin{figure}[ht!]
\begin{center}
\centerline{\includegraphics[width=1\columnwidth]{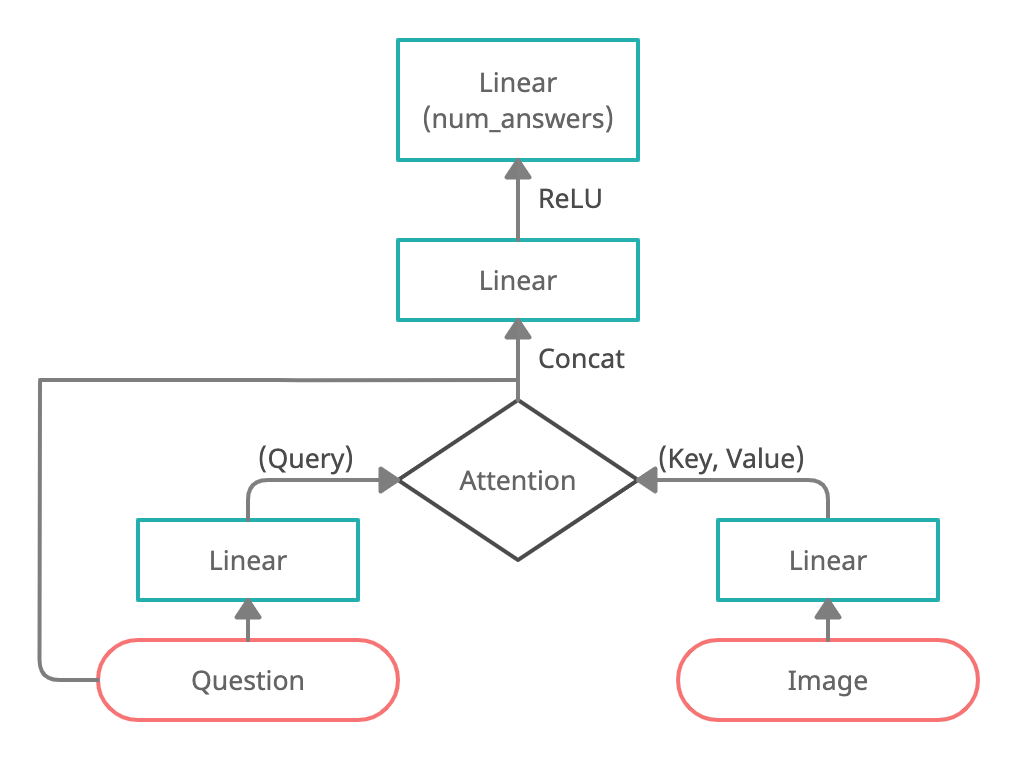}}
\caption{\textsc{attn} model architecture}
\label{fig:baseline_attn}
\end{center}
\vskip -0.2in
\end{figure}

In \textsc{attn}, we pass $B$ and $O$ through parallel fully-connected layers. We use the obtained question projection as the $query$ and the image projection as the $key$ \& $value$ in an attention mechanism to attend over the objects. Here, we use an attention mask to deal with the variable number of objects in each image. We concatenate the output of the attention mechanism to the question representation. This is followed by a fully-connected layer, ReLU and a classifying layer to get a distribution over the answers. The model architecture is illustrated in Figure  \ref{fig:baseline_attn}. Such attention-based techniques for multimodal-fusion have been used successfully in the past \citep{memexqa}.

\subsubsection{Unimodal Baselines}
For our unimodal baselines we remove the text component from the Concatenation Baseline to observe the impact of the image modality alone. Similarly, we remove the image component to observe the impact of the text modality alone.

\subsection{UNITER}
Given the state of the art performance achieved by the pre-trained \textsc{uniter} model on six different Vision + Language tasks, we use it to extract a cross-modal contextualized embedding $E$ for a given image and question. The embedding is passed through a multi-layer perceptron \textsc{(mlp)} which is trained during the process of fine-tuning \textsc{uniter}. The final layer of the \textsc{mlp} produces a softmax distribution over the answer set A, which is then used to predict an answer.

\subsection{Scene Graph Encoding}
\label{sec:graph_net}
Graph Neural Networks are a class of interpretable models successfully used for a range of tasks from relational reasoning to knowledge base completion \citep{zhou2018graph}. Following the work of \citep{gn_mac} we encode scene graphs using the class of graph neural networks introduced by \citet{battaglia2018relational} - Graph Networks \textsc{(GN)}. 

 As described in Equation \ref{eq:sg_def}, each node in the scene graph consists of a name and zero or more attributes. During pre-processing, we split the name and attributes into their constituent words, and represent a node with the vocabulary indices of these words. Similarly, an edge consists of the vocabulary indices of the relation name while the global vector is made up of the vocabulary indices of the question. These pre-processed representations are passed through a GloVe \citep{pennington-etal-2014-glove} embedding layer followed by a single-layer bi-directional LSTM network to obtain the input graph representation for the Graph Network, $(E_0, V_0, u_0)$. The Graph Network produces updated vectors following the algorithm described by \citep{zhou2018graph} which we show in Algorithm ~\ref{algo:gn}. Here, \textsc{mlp} refers to a multi-layer perceptron with two fully-connected layers of hidden size 512. 

 During iteration $t$, lines 2-4 update each edge using the value of the edge in iteration $t-1$, the node vectors for the source and receiver, and the global vector. Following the edge update, the nodes are updated in lines 5-9 using the updated values of incoming edges to the node, the previous node vector and the global vector. Finally, the global vector is updated (lines 10-14) using the updated values of all the nodes and edges. This process is repeated for 3 iterations with the output being updated edge, node and global vectors. 

\begin{algorithm}
\SetAlgoLined
\SetKwInput{kwPreCond}{Precondition}
\SetKwInput{kwPostCond}{Postcondition}
\kwPreCond{Initial Graph $(E_0, V_0, u_0)$}
\kwPostCond{Encoding $(E_T, V_t, u_T)$}
\For{$t \in \{1...T\}$}{
    \For{$k \in \{1...N^e\}$}{
        $e_{k}^{'}$ = \textsc{mlp}([$e_k, v_{r_k}, v_{s_k}, u$])
    }
    \For{$i \in \{1...N^v\}$}{
        let $E_{i}^{'}$ = ${(e_{k}^{'}, r_{k}, s_{k})}_{r_k=i,k=1:N^e}$ \\
        $\bar{e}_{i}^{'}$ = \textsc{mean}($E_{i}^{'}$) \\
        $v_{i}^{'}$ = \textsc{mlp}([$\bar{e}_{i}^{'}$, $v_i, u$]) \\
    }
    let $V^{'}$ = $\{v^{'}\}_{i=1:N^{v}}$ \\
    let $E^{'}$ = $\{(e_k^{'}, r_k, s_k)\}_{k=1:N^{e}}$ \\
    $\bar{e}^{'}$ = \textsc{mean}($E^{'}$) \\
    $\bar{v}^{'}$ = \textsc{mean}($V^{'}$) \\
    $u^{'}$ = \textsc{mlp}([$\bar{e}^{'}, \bar{v}^{'}, u$]) \\
    $(E, V, u)$ = $(E^{'}, V^{'}, u^{'})$
}
\KwRet{$(E, V, U)$}
\caption{Graph Network (\textsc{GN})} 
\label{algo:gn}
\end{algorithm}

\subsection{Question Answering with Scene Graphs}
As described in Section \ref{sec:graph_net}, the \textsc{gn} produces updated representations of the node, edge and global vectors. To convert these to a similar format as the object features $O$, we stack the node and global vectors to get the scene graph encoding, \textsc{sge}. The \textsc{attn} baseline (Section \ref{sec:baselines}) takes as input, $O$. We replace this with \textsc{sge} to get a question answering model \textsc{(sgatt)} that uses scene graphs. Similarly, we pass \textsc{sge} to \textsc{mac}, replacing the image input, to get \textsc{sgmac}. We jointly train the \textsc{gn} and \textsc{mac} from scratch, and use \textsc{gn} weights thus obtained, to train \textsc{sgatt}.

\subsection{Scene Graph Generation}
The problem of generating a scene graph from an image 
is typically split into 2 phases. The first phase consists of training an object detector 
such as Faster-RCNN. 
 The second phase uses the features extracted from this model to train a 
relation detector model responsible for identifying potential edge pairs between a set of objects and predicting the relationship between them. A popular work in the field of scene graph classification is the Neural Motifs \citep{zellers2018neural} model, which uses a stacked biLSTM structure to predict edge pairs and relations. 

In this work, we rely on a Faster R-CNN object feature extractor with a ResNet-50 backbone, paired with a Neural Motif model trained in 2 different variants (Section \ref{sec:labelloss}), to study the stability and variance associated with generating scene graphs.

\subsection{Training Curriculum}
\label{sec:train_curr}

\begin{figure*}[t!]
        \centering
        \includegraphics[scale=0.12]{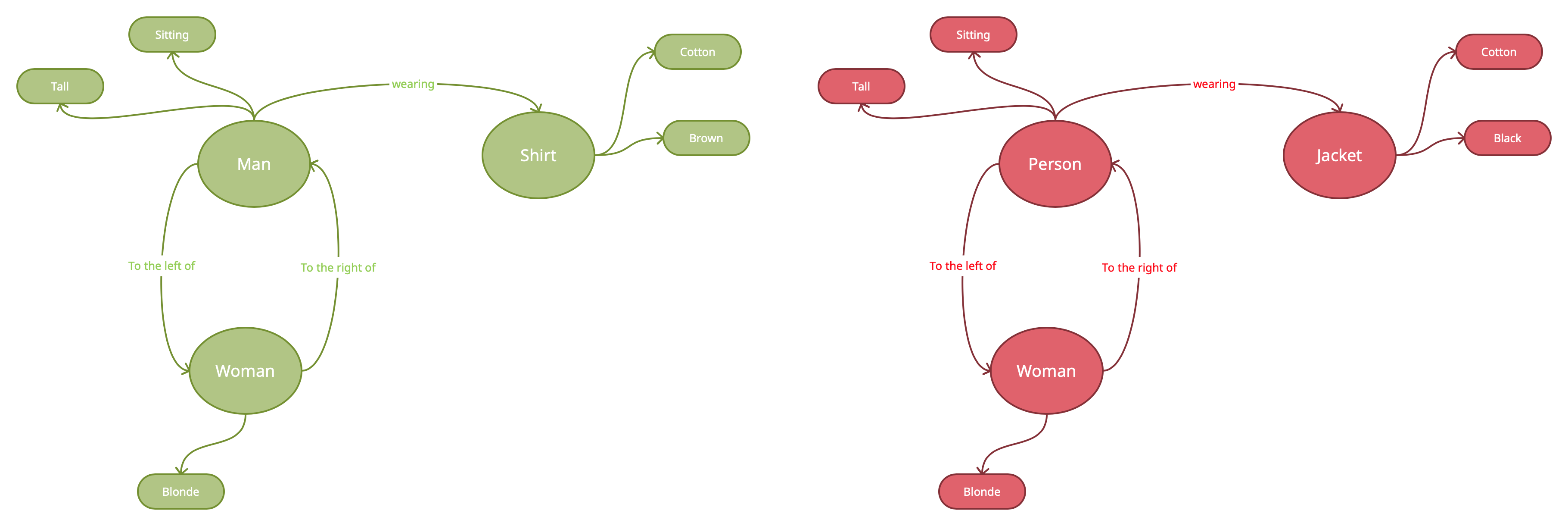}
        \caption{(Left) Ground truth scene graph and (Right) Generated noisy scene graph}
        \label{gt_noisy}
\end{figure*}
\begin{figure*}[t!]
    \centering
    \begin{subfigure}[t]{\columnwidth}
        \centering
        \includegraphics[scale=0.12]{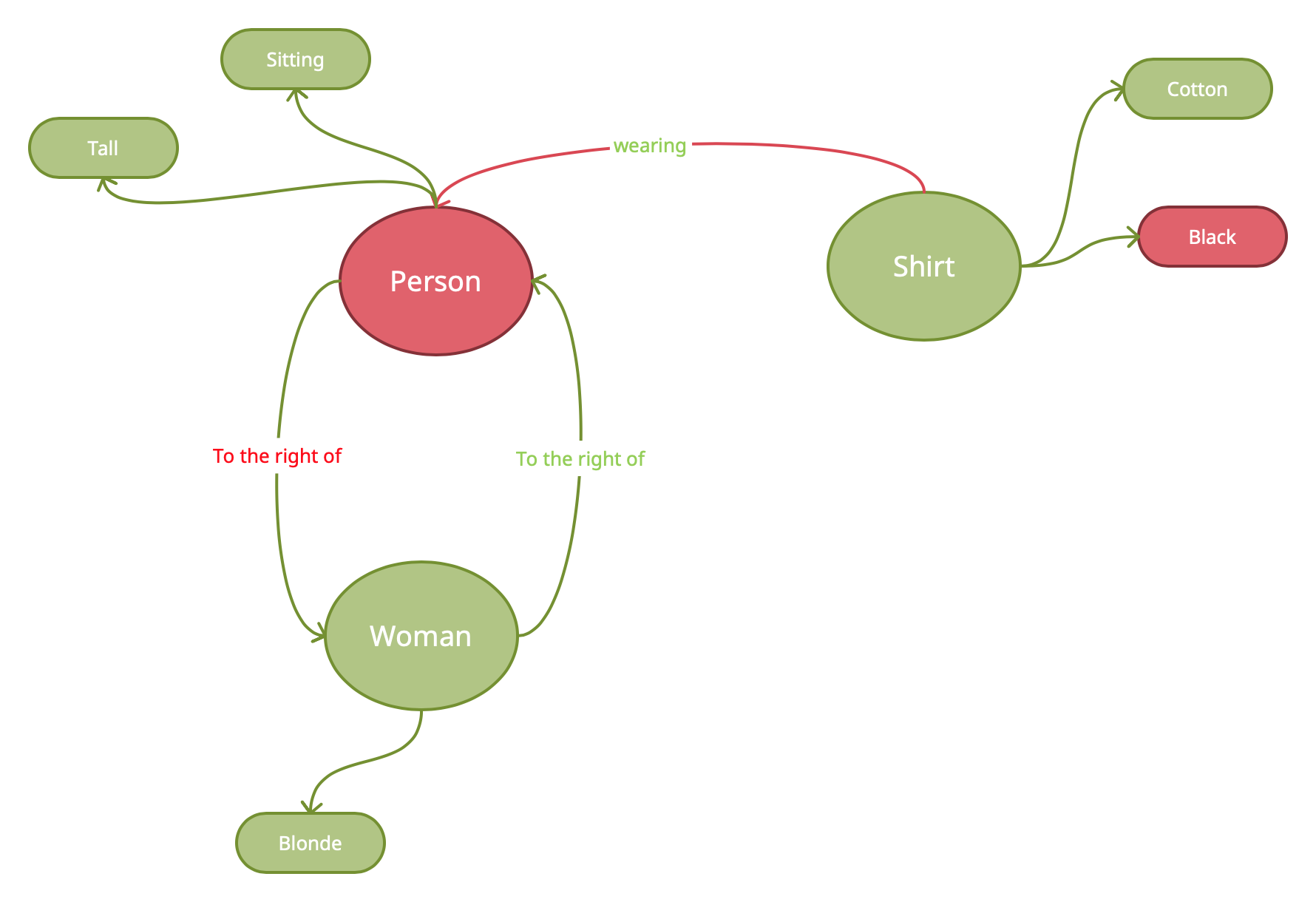}
        \caption{Scene graph generated after injecting noise}
        \label{noisy_injected}
    \end{subfigure}
    ~
    \begin{subfigure}[t]{\columnwidth}
        \centering
        \includegraphics[scale=0.12]{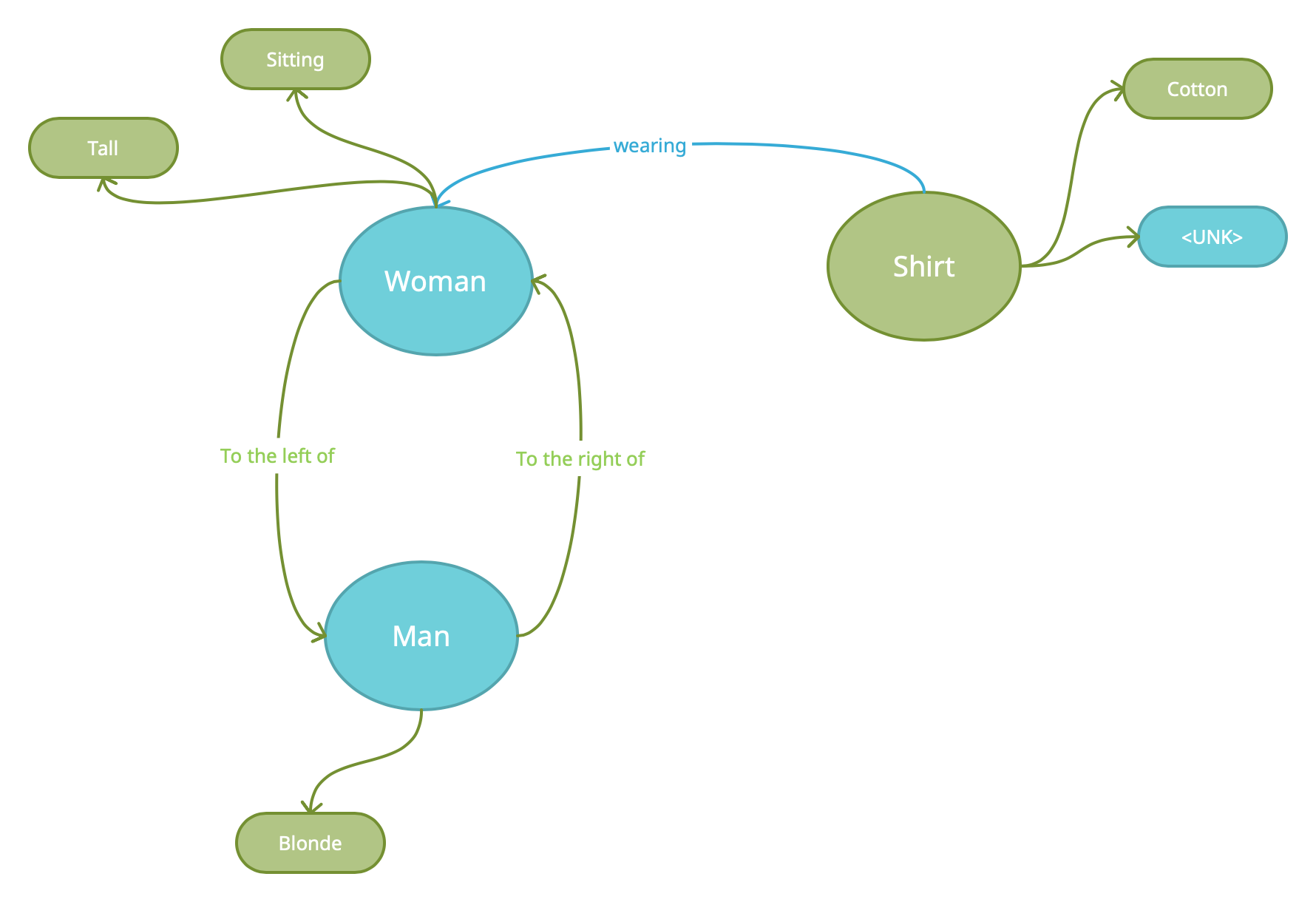}
        \caption{Scene graph generated after corrupting ground truth}
        \label{corrupt}
    \end{subfigure}
    \caption{Noisy/Corrupted Scene graph representations}
\end{figure*}

Since test images are evaluated on the noisy scene graphs, we propose a novel training curriculum to ensure the Graph Network representation is able to capture maximum information from the generated scene graphs. 
A Graph Network trained on only ground truth scene graphs encodes scene graphs for test images poorly since there is a large difference in the representations of ground truth and generated scene graphs due to differences in quality.

In order to train a model to have strong low-level features, we first train the Graph Network on ground truth scene graphs for 2 epochs and slowly introduce features of the noisy scene graphs into the ground truth scene graphs by randomly swapping out nodes, edge pairs and relations between the two (ground truth and noisy) graphs. The degree with which noise is introduced is controlled by 2 parameters: the \emph{probability} with which noise is introduced into the scene graph and the \emph{proportion} of the components for which we switch out with noisy counterparts. For example, if probability is 0.5 and proportion is 0.8 and we have 10 edges, then there is a 50\% chance that we switch out 8 ground truth edges with noisy edges. This injection of noise is slowly introduced while training the model with an increasing probability and proportion (of components) with each passing epoch. 

Maintaining a steady probabilistic approach of injecting noise into the ground truth scene graphs allows the model to have superior lower-level feature extraction as well as high-level semantic extraction in the upper layer more akin to the noisy graph features. Figures \ref{gt_noisy} and \ref{noisy_injected}  illustrate how a swap happens between components of the two scene graphs. We also experiment with switching out the entire graph (replace ground truth with noisy) at random during training, and with models that are trained on a mixed dataset.



\subsection{UNITER with Scene Graphs}
\begin{figure}[ht!]
\begin{center}
\centerline{\includegraphics[width=1\columnwidth]{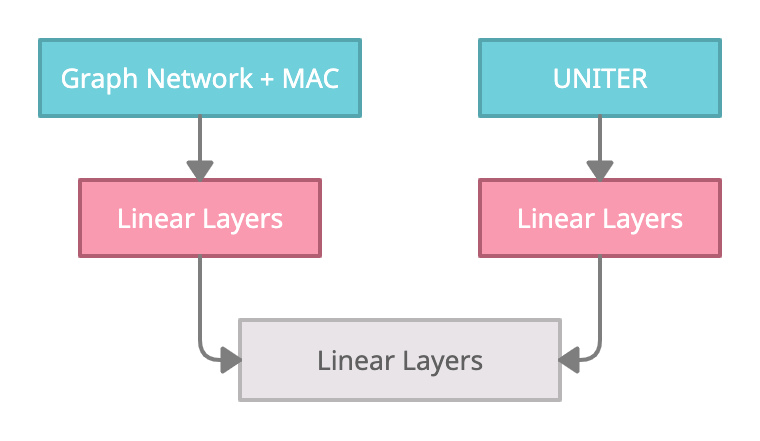}}
\caption{\textsc{uniter} with Scene Graphs}
\label{fig:uniter_sg}
\end{center}
\vskip -0.2in
\end{figure}

Given the success of pre-trained models across a wide range of tasks, we seek to use scene graphs with \textsc{uniter}. While the task-specific nature of \textsc{mac} allowed us to replace the image with the scene graph, doing so with \textsc{uniter} would not work since only image representations were fed to the model during pre-training. Hence, we adopt the late fusion architecture shown in Figure \ref{fig:uniter_sg}, where the output of an \textsc{sgmac} model trained on ground truth scene graphs is passed through fully-connected layers and the output of a pre-trained \textsc{uniter} model is passed through separate fully-connected layers. The outputs of these layers are then combined and passed through a classification layer. The fully-connected layers are trained from scratch, while \textsc{sgmac} and \textsc{uniter} are fine-tuned.

\section{Experimental Setup}
\label{sec:experimental_setup}
Throughout the course of this project, we primarily work with the GQA dataset, where we analyze the effects of using scene graphs instead of images to aid our models in the task of VQA. 

\subsection{Dataset}
The GQA dataset consists of 22M+ questions and 113K+ images, with associated scene graphs for 85K+ images. 
The dataset is divided into 5 splits: train, validation, testdev, test and challenge, with the option to avail a more balanced subset for most categories. The test and challenge datasets are used for leaderboard submissions and do not come with ground truth scene graphs. The training and validation sets are provided with images which have annotated ground truth scene graphs. Apart from the ground truth scene graphs which we leverage for our experiments, we also use the extracted object features and spatial features, provided with the dataset, in our baseline models.


The GQA task uses six metrics to evaluate a model's visual reasoning ability, namely \emph{accuracy}, \emph{validity}, \emph{plausibility}, \emph{consistency}, \emph{grounding}, and \emph{distribution}. Details about the calculation of each metric can be found in the GQA paper \citep{hudson2019gqa}.

\subsection{Training Environment}
The experimental setup for this project can be divided into various components based on how the training phase is carried out. The details regarding the various training stages and requirements are outlined in Table \ref{exp_table}.
\begin{table*}
\resizebox{\textwidth}{!}{
\begin{tabular}{|c|c|c|c|}
\hline
\textbf{Training phase}     & \textbf{Scene Graph Generation}    & \multicolumn{2}{c|}{\textbf{GQA Answering Model}} \\ \hline
\textbf{Model Architecture} & ResNet 50 + Neural Motifs & GN + \textsc{mac}            & \textsc{uniter}             \\ \hline
\textbf{Optimizer}          & SGD                       & Adam                & AdamW              \\ \hline
\textbf{Learning rate}      & 0.02                      & 1e-4                & 8e-5               \\ \hline
\textbf{Epochs / Steps}     & 50000 steps                     & 15 epochs                  & 6000 steps              \\ \hline
\textbf{Batch size}         & 4                         & 128                 & 500                \\ \hline
\textbf{No. of GPUs}        & 4                         & 1                   & 2                  \\ \hline
\textbf{GPU Model / RAM}    & Tesla V100 / 32 GB        & Tesla V100 / 16 GB  & Tesla V100 / 16 GB \\ \hline
\textbf{Time taken}         & 1 day 18 hours            & 22.5 hours          & 45 minutes         \\ \hline
\end{tabular}
}
\caption{Experimental settings of various models and stages of training}
\label{exp_table}
\end{table*}

\section{Results and Analysis}
\label{sec:results}
In this section, we present the results of our proposed approach along with an extensive analysis. We use the semantic types in \ref{sec:sem_appendix} for our analysis.
\subsection{Results of Baselines}

\begin{table}[ht]
\centering
\begin{tabular}{|c|c|c|c|}
    \hline
    \multicolumn{2}{|c|}{Our Models} & \multicolumn{2}{c|}{Comparable Models} \\
    \hline
    Model & Test & Model & Test \\
    \hline
    Image & 0.174 & CNN & 0.178 \\
    Text & 0.398 & LSTM & 0.41 \\
    \textsc{concat} & 0.435 & CNN+LSTM & 0.465 \\
    \textsc{attn} & 0.48 & Bottom-up & 0.497 \\
    \hline
\end{tabular}
\caption{Comparison of Baselines}
\label{tab:baseline_accuracy}
\end{table}

\begin{figure}[ht]
\begin{center}
\centerline{\includegraphics[width=1\columnwidth]{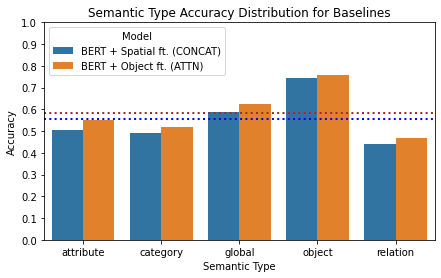}}
\caption{Comparison of the two baselines}
\label{fig:baseline_graph}
\end{center}
\vskip -0.2in
\end{figure}

\begin{table}[ht]
\centering
\begin{tabular}{|c|c|c|}
    \hline
    Model & Val & Testdev \\
    \hline
    \textsc{uniter} & 0.69 & 0.598 \\
    \textsc{sgmac} (GT) & 0.94 & - \\
    \textsc{sgmac} (Noisy) & 0.548 & 0.478 \\
    \textsc{sgmac} + \textsc{uniter} & 0.688 & 0.592 \\
    \hline
\end{tabular}
\caption{Results of Proposed Approaches}
\label{tab:proposed_approaches_accuracy}
\end{table}

From Table \ref{tab:baseline_accuracy} we see that the text only baseline does much better than the image only one. This is expected, as the question has more useful information to infer the answer than the image alone. The \textsc{attn} model achieves a significantly higher accuracy than \textsc{concat}. We believe that the attention mechanism helps \textsc{attn} focus on the most relevant objects in the image, based on the question at hand.

Figure \ref{fig:baseline_graph} shows the breakdown of the performance of both baselines on questions of different semantic types. \textsc{attn} does better than \textsc{concat} in every category. We see that there is a large scope for improvement in all the semantic types, especially in the relation type which primarily consists of questions that require knowledge about the relationships between objects to answer. This information can be found in the scene graphs.

\subsection{Performance of UNITER}
\begin{figure}[ht]
\begin{center}
\centerline{\includegraphics[width=1\columnwidth]{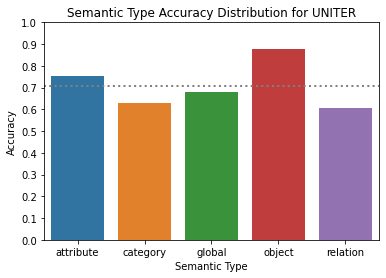}}
\caption{Performance of \textsc{uniter} over semantic types}
\label{fig:uniter_sem}
\end{center}
\vskip -0.2in
\end{figure}

Figure \ref{fig:uniter_sem} shows the performance of \textsc{uniter} over the different semantic types. The performance across all the types far exceeds that of the baselines, demonstrating the strength of \textsc{uniter}'s image-text representations and the advantages of pre-training. It is interesting to note that the relative ordering of semantic type accuracy is the same for \textsc{uniter} and the baselines. The accuracy on the object type is very high, while there is large room for improvement in the relation type. 
An analysis of the predictions of the model shows that for many examples, the model predicts an answer that is semantically close to the ground truth answer (e.g. 'bananas' vs 'banana', 'desk' vs 'table'). Here, it is useful to look to metrics other than accuracy which do not enforce an exact match with the ground truth answer. \textsc{uniter} obtains high plausibility (0.92) and validity (0.95) scores, indicating a good understanding of the image and text.

\subsection{\textsc{SGMAC} on GT Scene Graphs}
\begin{figure}[ht]
\begin{center}
\centerline{\includegraphics[width=1\columnwidth]{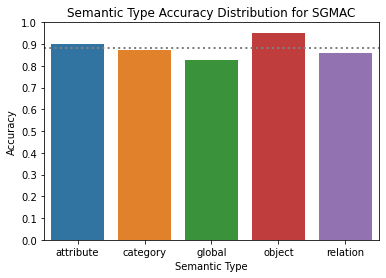}}
\caption{Performance of \textsc{sgmac} on ground truth scene graphs}
\label{fig:gn_sem}
\end{center}
\vskip -0.2in
\end{figure}

The performance of \textsc{sgmac} with GT scene graphs on different semantic types is shown in Figure \ref{fig:gn_sem}. This approach does very well over all types, and most notably the relation semantic type where \textsc{uniter} struggles. The overall validation accuracy of this approach is 94.6\%. This shows the effectiveness of the scene graph representation, and perhaps represents the upper limit of what can be achieved on GQA \footnote{\citet{gn_mac} report a score of 96.3 on training for 100 epochs, while we train for only 15-20 epochs.}. 

\subsubsection{\textsc{SGMAC} vs \textsc{SGATT}}
The attention model described in Section \ref{sec:baselines} with the Graph Network (66.4\% accuracy) does not match the performance of \textsc{sgmac} with the Graph Network (94.5\% accuracy), even when trained and evaluated on ground truth scene graphs. This demonstrates the importance of the manner in which the scene graph is parsed and proves that a good scene graph representation alone is not sufficient to achieve high accuracy.


\subsection{GT vs Generated Scene Graphs}
\begin{figure}[ht]
\begin{center}
\centerline{\includegraphics[width=1\columnwidth]{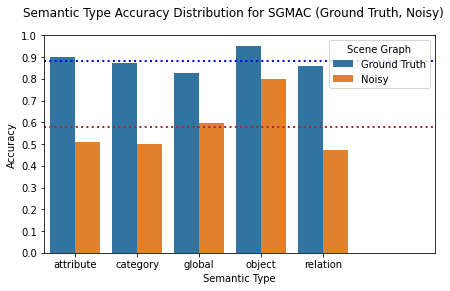}}
\caption{Comparison of \textsc{sgmac} using ground truth and noisy scene graphs}
\label{fig:noisy_gt_s}
\end{center}
\vskip -0.2in
\end{figure}

Figure \ref{fig:noisy_gt_s} shows a comparison of the semantic type accuracy of \textsc{sgmac} with GT scene graphs and noisy scene graphs.
 It is clear that the performance of \textsc{sgmac} with ground truth scene graphs far exceeds that of \textsc{sgmac} with noisy scene graphs on attribute, category and relation types and has a comparable performance on global and object types. We believe that this is due to the quality of the generated scene graphs rather than the technique used to encode them.

\begin{table}[ht!]
\centering
\begin{tabular}{|c|c|}
\hline
Scene Graphs & Val \\ \hline
GT & 0.945 \\
GT - No Relations & 0.851 \\
GT - No Attributes & 0.851 \\
Noisy & 0.529 \\
Noisy - No Relations & 0.532 \\
Noisy - No Attributes & 0.53 \\ \hline
\end{tabular}
\caption{\textsc{sgmac} performance on different scene graphs}
\label{tab:gn_mac_diff_sgs}
\end{table}

To get a better idea of the differences between the generated and ground truth scene graphs, we analyze their corresponding objects, attributes and relations.

\textbf{Objects}: On average, GT scene graphs have 16 objects, and 3 of these are not present in the generated scene graphs. Since most objects are correctly detected, the model correctly answers the majority of questions about the existence of objects in the image (object semantic type). The performance on this type using GT and generated scene graphs is comparable due to the high overlap in the objects. 

\textbf{Attributes}: We find that most attributes in the GT scene graphs are present in the generated scene graphs. However, the generated scene graphs additionally contain many spurious attributes, which negates the effect of getting the attributes right and explains the poor performance on the attribute semantic type. For example, a phone which is grey in color might also have ``blue" and ``brown" as additional attributes, and hence a question such as ``What color is the phone?" would not be answered correctly. In Sec. \ref{sec:filtering_sg} we describe the use of thresholds to reduce spurious attributes.

\textbf{Relations}: We find that most of the edges have extremely low overlap (1-2 edges) between GT and noisy scene graphs, which explains why the gap between the performance of the two on the relation semantic type is high.

The above analysis showcases the need for improvement in the quality of the generated scene graphs. Specifically, correctly detecting relations would go a long way towards improving performance. It also shows how the use of accurate scene graphs would benefit VQA.

In addition to qualitative analysis, we conduct the ablation studies shown in Table \ref{tab:gn_mac_diff_sgs} to understand the impact of various components of scene graphs. We run experiments wherein we remove either relation names or attributes from the scene graph. On removing relations/attributes from GT scene graphs, we observe a 9.4\% drop in accuracy. The same experiments when run on generated scene graphs yield little difference in accuracy further indicating that the quality of detected relations is poor. 

\subsection{Effect of scene graph quality}
We further experiment with various scene graphs outputs that were generated by different variations of the Neural Motifs model, to effectively analyze the difference in quality of the scene graphs generated. We experiment with 2 variations of the model, making changes to the loss function to better understand the differences in prediction pertaining to the predicted labels for an edge relation. The variations are as follows:
\subsubsection{Impact of Label Loss Smoothing}
\label{sec:labelloss}
 During a preliminary analysis of the dataset, we noticed that there is a stark imbalance in the training data for the relation labels ``to the left of" and ``to the right of". Both labels appear at least 18 times more often than next most frequent relation in the training set. This causes the standard variant of our model to disproportionately predict the above mentioned relations. In an attempt to mitigate this class imbalance, we use an exponential label smoothing loss function.
 
\begin{table}[ht!]
\centering
\begin{tabular}{|c|c|c|}
\hline
Model & Testdev \\ \hline
Without Smoothing - Noisy & 0.458 \\
Without Smoothing - GT + Noisy & 0.479 \\
With Smoothing - Noisy & 0.457 \\
With Smoothing - GT + Noisy & 0.476  \\ \hline
\end{tabular}
\caption{\textsc{sgmac} performance with and without label smoothing}
\label{tab:gn_label_smoothing}
\end{table}

We can see from Table \ref{tab:gn_label_smoothing} that the label smoothing does not help improve the downstream model performance much. Another potential remedy to handle the class imbalance would be to use a two-way softmax that separates the ``to the left of " and ``to the right of" labels from the other relations.

\subsubsection{Corruption of GT Scene Graphs}

\begin{table}[ht!]
\centering
\begin{tabular}{|c|c|c|}
\hline
Level of Corruption & Val \\ \hline
0.2 & 0.861 \\
0.4 & 0.788 \\
0.6 & 0.721 \\
0.8 & 0.662  \\ \hline
Noisy & 0.461 \\ \hline
\end{tabular}
\caption{\textsc{sgmac} performance with different levels of scene graph corruption}
\label{tab:gn_mac_corrupt}
\end{table}

In order to understand how the noisy scene graphs affect downstream performance in our GQA model, we conduct evaluation of our best performing GT (\textsc{sgmac}) model on varying levels of corrupted ground truth scene graphs. This allows us to estimate the amount of noise it would take in a gold standard scene graph dataset to reach the same accuracy we obtain when evaluated on a noisy scene graph dataset. In order to corrupt the ground truth scene graphs, we randomly swap nodes, edge pairs and relations, and also corrupt object attributes and names by replacing them with \emph{UNK} tokens.  Figure \ref{corrupt}  illustrates an example of what a corrupted ground truth scene graph would look like, in contrast to the ground truth scene graph, as depicted on the left side of Figure \ref{gt_noisy}. These differ from the noisy injected scene graphs (Section \ref{sec:train_curr}) as we do not bring in any external components from the noisy scene graphs into the corrupted GT scene graph. We create 4 validation datasets of different levels of corruption by varying degrees of probability with which we corrupt the scene graph as shown in Table \ref{tab:gn_mac_corrupt}. From the table it is clear that the noisy scene graphs differ much more significantly from the ground truth representations and that this level of noisiness will be very hard to overcome even with the most powerful model capable of reasoning over scene graphs. 

\subsubsection{Filtering Predicted Scene Graphs}
\label{sec:filtering_sg}
In order to remove any redundant or spurious information that is included while generating the noisy scene graphs, we also employ some standard filtering techniques to take the top k components of the noisy scene graph. In an unfiltered noisy scene graph, we would have 80 objects, 10 attributes/object and 100 relations between object pairs. From this set, we employ two filtering strategies based on both the confidence score of the model for each label and also based on the top k labels the model predicts. \par 
For the top k setting, we take the top 40 objects, the top 3 attributes/object and the top 80 relations from each noisy scene graph. We also alternatively filter them by a threshold value and only include components that are greater than a particular threshold. The threshold values selected are 0.1 for objects, 0.9 for attributes and 0.8 for relations.

\begin{table}[ht!]
\centering
\begin{tabular}{|c|c|}
\hline
Scene Graph Filter & Testdev \\ \hline
No Filter - Noisy & 0.468 \\
Top k - Noisy & 0.464 \\
Threshold - Noisy & 0.458 \\ \hline
No Filter - GT + Noisy & 0.475 \\
Top k - GT + Noisy & 0.478 \\
Threshold - GT + Noisy & 0.479 \\ \hline
\end{tabular}
\caption{\textsc{sgmac} performance with different filters on scene graphs}
\label{tab:mac_gn_filter}
\end{table}

When training from scratch, retaining all the model attributes and relations intact seems to let the model learn more information with further training. This indicates that the filtered noisy scene graphs are possibly more akin to the ground truth representations but without a good lower level feature representation (a good starting checkpoint), the Graph Network might not be powerful enough to learn much from the (sparsely) filtered noisy scene graphs.

\subsection{Training Curriculum}
\label{tc_result}
\begin{table}[ht!]
\centering
\begin{tabular}{|c|c|}
\hline
Training Curriculum & Testdev \\ \hline
GT & 0.373 \\
Noisy & 0.468 \\
GT+Noisy & 0.475 \\
Probabilistic & 0.471 \\
Probabilistic (Filtered) & 0.477 \\
Probabilistic (Complete) & 0.478 \\ \hline
\end{tabular}
\caption{Training Curriculum Results}
\label{tab:training_curriculum}
\end{table}
As mentioned in Section \ref{sec:train_curr}, we see that training with just the GT and or just noisy scene graphs does not yield very desirable results. However, training the model using the novel training curriculum outlined in Section \ref{sec:train_curr} gives us a gain in performance.

We see that gradually swapping ground truth scene graphs with noisy scene graphs during training (Probabilistic (Complete) in \ref{tab:training_curriculum}) performs better than completely replacing ground truth scene graphs with noisy scene graphs after a few epochs (GT + Noisy in \ref{tab:training_curriculum}). 
Completely replacing the ground truth scene graphs (for random noisy graphs) performs on par with the probabilistic training mechanism explained in Figure \ref{gt_noisy} and \ref{noisy_injected}.

We see that filtering the noisy dataset on the top k objects/attributes/relations helps weed out the redundant/spurious labels and improve accuracy slightly.  Overall, these results prove that the probabilistic training curriculum helps the model generalize much better compared to training on one source of data, and hence validate the effectiveness of the proposed novel training curriculum.


\subsection{UNITER with Scene Graphs}
The late-fusion ensemble of \textsc{uniter} with \textsc{sgmac} does not perform better than \textsc{uniter} itself. We analyse the accuracies across semantic types to see if it does better on relational queries, but find no significant differences. It could be that \textsc{uniter} is a very powerful model which overwhelms the much smaller \textsc{sgmac}.

\section{Conclusion and Future Work}
\label{sec:contributions}
In this work, we present novel research that is aimed at understanding various aspects of the use of scene graphs for the task of VQA. We develop modules for scene graph encoding, architectures for question answering using scene graphs, probabilistic training curricula, and late fusion models for question answering. We primarily use validation and test-dev scores for our analysis, but our test submissions can be found on the leaderboard\footnote{\url{https://eval.ai/web/challenges/challenge-page/225/leaderboard/733}} under the team name ``bert".

The architecture combining a Graph Network with \textsc{mac} surpasses the performance of state of the art models when ground truth scene graphs are available. However, this performance quickly deteriorates as reliance on auto-generated scene graphs increases. We show that a training curriculum that utilizes generated and ground truth scene graphs is more effective than the use of a singular type of scene graph. This approach does not come close to matching pre-trained image based models such as \textsc{uniter}, suggesting that improving scene graph generation will be the most important step for using scene graph based models for VQA. Tackling the class imbalance for relations and improving the object detection could help in this regard. Techniques that allow the model to fall-back to image features based on the level of noise in the scene graph could be quite effective and we plan to explore these further.

\section*{Acknowledgments}
We would like to express our gratitude to Dr. Lu Jiang (Google) for providing us invaluable advice and guidance throughout the course of this work. We would also like to thank Dr. Mayu Otani for her continued support.

\bibliography{acl2019}
\bibliographystyle{acl_natbib}

\appendix

\section{Appendices}
\label{sec:appendix}
\subsection{Semantic Types in GQA}
\label{sec:sem_appendix}
GQA classifies its questions into 5 semantic types:
\begin{enumerate}[noitemsep]
    \item \textbf{Relation}: subject/object of a described relation, e.g. “What furniture is the beer can on?”
    \item \textbf{Attribute}: properties/position of an object, e.g. “What is the color of the man’s shirt?”
    \item \textbf{Object}: existence of objects in the image, e.g. “Are there any wine glasses or cups?”
    \item \textbf{Category}: object identification within a class, e.g. “Which kind of clothing is maroon?”
    \item \textbf{Global}: overall scene properties, e.g. “How's the weather?”
\end{enumerate}

\end{document}